\pdfoutput=1

\documentclass[11pt]{article}

\usepackage[]{acl}

\usepackage{times}
\usepackage{latexsym}

\usepackage[T1]{fontenc}

\usepackage[utf8]{inputenc}

\usepackage{microtype}
\usepackage{amsmath}
\usepackage{graphics} 
\usepackage{graphicx} 
\usepackage{multirow}
\usepackage{subcaption}
\usepackage{lettrine}
\usepackage{array}

\usepackage{enumitem} 
\setlist[itemize]{noitemsep, topsep=0pt}
\setlist[enumerate]{noitemsep, topsep=0pt}

\usepackage{xspace}
\newcommand{\LM}{\textsc{LM}\xspace}
\newcommand{\MT}{\textsc{MT}\xspace}

\usepackage{url}
\title{Revisiting Syllables in Language Modelling and their \\Application on Low-Resource Machine Translation}

\author{Arturo Oncevay${ }^{\epsilon,\chi}$ ~ \quad Kervy Dante Rivas Rojas${ }^{\rho,\alpha}$ \quad \\ \textbf{Liz Karen Chavez Sanchez${ }^{\chi}$ \quad Roberto Zariquiey${ }^{\rho,\chi}$} \\
${ }^{\epsilon}$School of Informatics, University of Edinburgh, Scotland \\ ${ }^{\rho}$Pontificia Universidad Católica del Perú (${ }^{\alpha}$IA-PUCP $|$ ${ }^{\chi}$Chana Field Station), Peru \\
  \texttt{a.oncevay@ed.ac.uk,rzariquiey@pucp.edu.pe} \\}

\begin{document}
\maketitle
\begin{abstract}
Language modelling and machine translation tasks mostly use subword or character inputs, but syllables are seldom used. Syllables provide shorter sequences than characters, require less-specialised extracting rules than morphemes, and their segmentation is not impacted by the corpus size. 
In this study, we first explore the potential of syllables for open-vocabulary language modelling in 21 languages. We use rule-based syllabification methods for six languages and address the rest with hyphenation, which works as a syllabification proxy. With a comparable perplexity, we show that syllables outperform characters and other subwords. Moreover, we study the importance of syllables on neural machine translation for a non-related and low-resource language-pair (Spanish--Shipibo-Konibo). In pairwise and multilingual systems, syllables outperform unsupervised subwords, and further morphological segmentation methods, when translating into a highly synthetic language with a transparent orthography (Shipibo-Konibo). Finally, we perform some human evaluation, and discuss limitations and opportunities.
\end{abstract}

\section{Introduction}
In language modelling (\textsc{LM}), we learn distributions over sequences of words, subwords or characters, and the last two can allow an open-vocabulary generation~\cite{Sutskever:2011:GTR:3104482.3104610}. We rely on subword segmentation as a widespread approach to generate rare subword units \cite{sennrich-etal-2016-neural}. However, the lack of a representative corpus, in terms of the word vocabulary, can constrain the unsupervised segmentation (e.g. with scarce monolingual texts \cite{joshi-etal-2020-state}). 
As an alternative, we could use character-level modelling, since it also has access to subword information \cite{kim-etal-2016-character}, 
but we face long-term dependency issues and require longer training time to converge. Similar issues are extended to other generation tasks, such as machine translation (\textsc{MT}).

In this context, 
we focus on syllables, which are speech units: ``A syl-la-ble con-tains a sin-gle vow-el u-nit''. syllables can be defined as a group of segments that is pronounced as a single articulatory movement. Syllables are fundamental phonological units since they participate in important word prosodic patterns, such as stress assignment. In this sense, syllables are more linguistically relevant units than characters, and behave as a mapping function to reduce the length of the sequence with a larger ``alphabet'' or syllabary. Their extraction can be rule-based and corpus-independent, but data-driven methods or hyphenation using dictionaries can approximate them as well. 

We assess whether syllables are useful for encoding and/or decoding a diverse set of languages on two generation tasks. First, for \LM, we study 21 languages, to cover different levels of orthographic depth, which is the degree of grapheme-phoneme correspondence \cite{borgwaldt-etal-2005-onset} and a factor that can increase complexity to syllabification \cite{marjou-2021-oteann}.\footnote{E.g., English has a deep orthography (weak correspondence), whereas Finnish is transparent \cite{ziegler-etal-2010-orthographic}.} Whereas for \MT, we focus on the distant and low-resource language-pair of Spanish--Shipibo-Konibo. We choose Shipibo-Konibo\footnote{See Appendix \ref{app:shipibo} for more details about the language.} because it is an endangered language with scarce textual corpora, which limits unsupervised segmentation methods, and has a transparent orthography, which could be beneficial to syllabification. Also, we consider multilingual MT systems, as they outperformed pairwise systems for the chosen language pair \cite{mager-etal-2021-findings}.




\section{Related work}
\label{sec:related-work}


The closest \LM study to ours is from \citet{mikolov-etal-2012-subword} for subword-grained prediction in English, where they used syllables as a proxy to split words with low frequency, reduce the vocabulary and compress the model size. Besides, syllable-aware \LM was addressed by \citet{assylbekov-etal-2017-syllable} for English, German, French, Czech, Spanish and Russian, and by \citet{yu-etal-2017-syllable} for Korean. However, in both cases, the syllables were composed with convolutional filters into word-level representations for closed-vocabulary generation. Besides, for subword-aware open-vocabulary \LM, \citet{blevins-zettlemoyer-2019-better} incorporated morphological supervision with a multi-task objective.

For syllable-based \MT, there are mostly studies for related paired languages, such as Indic languages  (in statistical \MT without subword-based baselines:  \citet{kunchukuttan-bhattacharyya-2016-orthographic}), Tibetan--Chinese \cite{lai-etal-2018-tibetan}, and Myanmar--Rakhine \cite{myint-oo-etal-2019-neural}. Instead, Spanish--Shipibo-Konibo is a non-related language-pair. The only distant pair was English--Myanmar \cite{shwesin-etal-2019-ucsynlp}, but they did not compare it with unsupervised subwords. Neither of these studies analysed multilingual settings.  

\section{Open-vocabulary language modelling with a comparable perplexity}


\paragraph{Open-vocabulary output} We generate the same input unit (e.g. characters, syllables or other subwords) as an open-vocabulary \LM task, where there is no prediction of an ``unknown'' or out-of-vocabulary word-level token \cite{Sutskever:2011:GTR:3104482.3104610}. 
We thereby differ from previous works, and refrain from composing the syllable representations into words to evaluate only word-level perplexity. 

\paragraph{Character-level perplexity} For a fair comparison across all granularities, we evaluate all results with character-level perplexity: 
\begin{equation}
\operatorname{ppl}^{c} = \exp{(\mathcal{L}_{\mathrm{LM}}(\mathbf{s})\cdot\frac{|\mathbf{s}^{\text{seg}}|+1}{|\mathbf{s}^{c}|+1})}
\end{equation}

where $\mathcal{L}_{\mathrm{LM}}(\mathbf{s})$ is the cross entropy of a string $\mathbf{s}$ computed by the neural \LM, and $|\mathbf{s}^{\text{seg}}|$ and $|\mathbf{s}^{c}|$ refer to the length of $\mathbf{s}$ in the chosen segmentation and character-level units, respectively \cite{miekel2019charppl}. The extra unit considers the end of the sequence.

\subsection{Experimental setup}

\paragraph{Languages and datasets}
Corpora are listed in Table \ref{tab:data-splits} in Appendix \ref{app:datasets}. 
We first choose \mbox{WikiText-2-raw~\cite[en\textsubscript{w};][]{merity2016pointer}}, which contains around two million word-level tokens extracted from Wikipedia articles in English. Furthermore, 
we employ 20 Universal Dependencies \cite[UD;][]{nivre-etal-2020-universal} treebanks, similarly to \citet{blevins-zettlemoyer-2019-better}.\footnote{The languages are chosen given the availability of an open-source syllabification or hyphenation tool. We prefer to use the UD treebanks, instead of other well-known datasets for language modelling (e.g. Multilingual Wikipedia Corpus \cite{kawakami-etal-2017-learning}), because they provide morphological annotations, which are fundamental for this study.} Finally, we include the Shipibo-Konibo (shp) side of the parallel corpora provided by the AmericasNLP shared task on \MT \cite{mager-etal-2021-findings}, which is also used in \S\ref{sec:mt}.

\paragraph{Syllable segmentation (\textsc{Syl})} For splitting syllables in different languages, we used rule-based syllabification tools for English, Spanish, Russian, Finnish, Turkish and Shipibo-Konibo, and dictionary-based hyphenation tools for all languages except the ones mentioned above. We list the tools in Appendix \ref{app:segmentation}. 

\paragraph{Segmentation baselines} Besides characters (\textsc{Char}) and the annotated morphemes in the UD treebanks (\textsc{Morph}), we consider Polyglot (\textsc{Poly})\footnote{\url{polyglot-nlp.com}}, which includes models for unsupervised morpheme segmentation trained with Morfessor \cite{virpioja-2013-morfessor}. Moreover, we employ an unsupervised subword segmentation baseline of Byte Pair Encoding~\cite[BPE;][]{sennrich-etal-2016-neural}\footnote{We use: \url{https://github.com/huggingface/tokenizers}} with different vocabulary sizes from 2,500 to 10,000 tokens, with 2,500 steps. We also fix the parameter to the syllabary size%
. Appendix \ref{app:segmentation} includes details about the segmentation format.

\paragraph{Model and training} Following other open-vocabulary \LM studies~\cite{mielke-eisner-2019-spell, mielke-etal-2019-kind}, we use a low-compute version of an LSTM neural network, named Average SGD Weight-Dropped~\cite{merity2018regularizing}. 
See the hyperparameter details in Appendix \ref{app:model}.

\begin{table}
    \centering
    \setlength\tabcolsep{2.5pt}
    \resizebox{\linewidth}{!}{%
    \begin{tabular}{l|ccccc}
 & \textsc{Char} & \textsc{Morph} & \textsc{Poly} & \textsc{Syl} & \textsc{BPE}\textsubscript{best} \\ \hline \hline
 en\textsubscript{w}* & 2.48 \small{$\pm$0.0} & - & 2.8 \small{$\pm$0.0} & \textbf{1.96} \small{$\pm$0.0} & 2.91 \small{$\pm$0.0} \\ \hline
bg & 3.56 \small{$\pm$0.03} & 4.09 \small{$\pm$0.05} & 4.69 \small{$\pm$0.01} & \textbf{2.87} \small{$\pm$0.0} & 5.19 \small{$\pm$0.01} \\
ca & 2.84 \small{$\pm$0.0} & 3.11 \small{$\pm$0.02} & 3.26 \small{$\pm$0.01} & \textbf{2.21} \small{$\pm$0.0} & 3.31 \small{$\pm$0.0} \\
cs & 3.32 \small{$\pm$0.0} & 3.11 \small{$\pm$0.01} & 4.18 \small{$\pm$0.01} & \textbf{2.66} \small{$\pm$0.0} & 4.24 \small{$\pm$0.0} \\
da & 4.25 \small{$\pm$0.01} & 4.42 \small{$\pm$0.04} & 5.6 \small{$\pm$0.0} & \textbf{3.1} \small{$\pm$0.01} & 6.21 \small{$\pm$0.03} \\
de & 3.5 \small{$\pm$0.04} & 3.36 \small{$\pm$0.08} & 3.79 \small{$\pm$0.0} & \textbf{2.48} \small{$\pm$0.0} & 3.86 \small{$\pm$0.02} \\
en\textsubscript{}* & 4.11 \small{$\pm$0.01} & 4.39 \small{$\pm$0.08} & 5.67 \small{$\pm$0.01} & \textbf{2.82} \small{$\pm$0.07} & 5.65 \small{$\pm$0.04} \\
es* & 3.16 \small{$\pm$0.01} & 3.71 \small{$\pm$0.04} & 3.95 \small{$\pm$0.01} & \textbf{2.51} \small{$\pm$0.0} & 3.98 \small{$\pm$0.0} \\
fi* & 3.77 \small{$\pm$0.01} & 4.05 \small{$\pm$0.12} & 4.76 \small{$\pm$0.01} & \textbf{3.1} \small{$\pm$0.0} & 5.27 \small{$\pm$0.01} \\
fr & 3.09 \small{$\pm$0.01} & 3.67 \small{$\pm$0.02} & 3.82 \small{$\pm$0.01} & \textbf{2.3} \small{$\pm$0.01} & 3.87 \small{$\pm$0.01} \\
hr & 3.52 \small{$\pm$0.02} & 3.92 \small{$\pm$0.01} & 4.34 \small{$\pm$0.0} & \textbf{2.8} \small{$\pm$0.0} & 4.52 \small{$\pm$0.02} \\
it & 2.8 \small{$\pm$0.0} & 3.19 \small{$\pm$0.0} & 3.43 \small{$\pm$0.01} & \textbf{2.27} \small{$\pm$0.01} & 3.61 \small{$\pm$0.0} \\
lv & 4.55 \small{$\pm$0.02} & 5.31 \small{$\pm$0.0} & 6.82 \small{$\pm$0.02} & \textbf{3.59} \small{$\pm$0.0} & 7.19 \small{$\pm$0.0} \\
nl & 3.83 \small{$\pm$0.05} & 3.69 \small{$\pm$0.1} & 4.44 \small{$\pm$0.01} & \textbf{2.76} \small{$\pm$0.01} & 4.83 \small{$\pm$0.01} \\
pl & 4.03 \small{$\pm$0.01} & 4.77 \small{$\pm$0.22} & 5.96 \small{$\pm$0.04} & \textbf{3.19} \small{$\pm$0.0} & 5.99 \small{$\pm$0.0} \\
pt & 3.31 \small{$\pm$0.01} & 3.46 \small{$\pm$0.03} & 4.03 \small{$\pm$0.01} & \textbf{2.56} \small{$\pm$0.0} & 4.24 \small{$\pm$0.01} \\
ro & 3.4 \small{$\pm$0.02} & 3.89 \small{$\pm$0.04} & 4.25 \small{$\pm$0.01} & \textbf{2.72} \small{$\pm$0.0} & 4.71 \small{$\pm$0.01} \\
ru* & 3.28 \small{$\pm$0.0} & 2.93 \small{$\pm$0.01} & 4.05 \small{$\pm$0.0} & \textbf{2.69} \small{$\pm$0.01} & 4.04 \small{$\pm$0.0} \\
sk & 6.16 \small{$\pm$0.05} & 5.1 \small{$\pm$0.07} & 7.61 \small{$\pm$0.08} & \textbf{4.62} \small{$\pm$0.01} & 10.51 \small{$\pm$0.03} \\
tr* & 4.16 \small{$\pm$0.05} & 4.86 \small{$\pm$0.05} & 6.41 \small{$\pm$0.07} & \textbf{3.66} \small{$\pm$0.03} & 6.98 \small{$\pm$0.1} \\
uk & 4.92 \small{$\pm$0.02} & 6.45 \small{$\pm$0.11} & 8.11 \small{$\pm$0.03} & \textbf{4.24} \small{$\pm$0.02} & 9.23 \small{$\pm$0.02} \\  \hline
shp* & 4.48\small{$\pm$0.01} & - & - &  \textbf{2.15}\small{$\pm$0.02} &  3.50\small{$\pm$0.03}\\ \hline

    \end{tabular}
    }
    \caption{Character-level perplexity ($\downarrow$) in test. We show the mean and standard deviation for three runs with different seeds. \textsc{BPE} shows the best result from models with different vocabulary sizes. \textsc{Syl} presents the syllabification-based result if it is available (*), or the hyphenation otherwise.}
    \label{tab:perplexity}
\end{table}

\subsection{Results and discussion}

Table \ref{tab:perplexity} shows the $\operatorname{ppl}^c$ values for the different levels of segmentation we considered in the study, where we did not tune the neural \LM model for a specific segmentation. We observe that syllables always result in better perplexities than other granularities, even for deep orthography languages such as English or French. 
The results obtained by the BPE baselines are relatively poor as well, and they could not beat characters in any dataset, 
even though we searched for an optimal vocabulary size for the BPE algorithm. The advantage of using syllables is that we do not need to tune a hyper-parameter to extract a different set of subword pieces.


As a significant outcome, we note that syllables did not fail to beat characters, at least in an open-vocabulary \LM task, which extends the findings of \citet{assylbekov-etal-2017-syllable}. One potential reason is that character-level modelling requires a larger model capacity due to the longer sequences, however, that is also an advantage towards syllables. Besides, other subword pieces with a closer sequence length to syllables (BPE, \textsc{Morph} or \textsc{Poly}) were still outperformed.

Finally, in Appendix \ref{app:discusssion}, we further discuss the relationship between the syllable type/token ratio with the word vocabulary growth and perplexity. 




\section{Low-resource Machine Translation}
\label{sec:mt}

After observing the positive impact on \LM, we focus on syllables for \MT, which adds complexity to the process, as there is at least one extra language involved. In contrast to prior work, we (i) study syllable-based \MT for a distant and low-resource language-pair, Spanish--Shipibo-Konibo; (ii) compare syllables against the most widespread unsupervised segmentation method (BPE) with automatic metrics and human evaluation; and (iii) analyse the applicability of syllables on multilingual translation systems. The last element is significant, as a multilingual setting is the state-of-the-art approach for leveraging low-resource language-pairs performance \cite{siddhant-etal-2022-towards}. Moreover, we decided to apply syllabification only on Shipibo-Konibo, a highly synthetic\footnote{With a high ratio of number of morphemes per word.} language with scarce textual data and with a transparent orthography\footnote{We attempted to use syllables on Spanish and English as well, but with negative results. With large data, unsupervised segmentation methods like BPE can obtain more significant and overlapping subwords from source and target.}.

For this reason, we focus in three settings. First, \textsc{mono}, a pairwise system where each source and target is segmented with a different method. Second, \textsc{joint}, another pairwise system where the BPE baseline is jointly trained with the source and target data. Third, \textsc{o2m}, a multilingual one-to-many\footnote{We do not consider the many-to-one direction due to resource constraints, and because we observed that the improvements by syllables are noted when decoding Shipibo-Konibo.} system, where the BPE baseline is jointly trained with all the languages (we added Spanish--English in our experiments). 

\subsection{Experimental setup}

\paragraph{Data} For Spanish--Shipibo-Konibo (es--shp), we use the dataset provided by the AmericasNLP workshop (\citet{mager-etal-2021-findings,galarreta-etal-2017-corpus,montoya-etal-2019-continuous}), and perform the same split as \citet{mager-etal-2022-bpe} for the dev and test subsets, to make the results comparable to their morphological segmentation experiments. For the multilingual case, we use the Spanish--English (es--en) train set from EuroParl \cite{koehn-2005-europarl} and newscommentary-v8, and the \textsc{newstest2013.es-en} \cite{bojar-etal-2013-findings} evaluation sets.

\paragraph{Segmentation} (i) \textsc{BPE} \cite{sennrich-etal-2016-neural} is our baseline segmentation method, and we use the implementation of SentencePiece \cite{kudo-richardson-2018-sentencepiece}. Similar to \citet{mager-etal-2022-bpe}, we fix the best vocabulary size at 5000 pieces for the \textsc{mono} setting, after trying different values from 1k to 10k. \textsc{joint} and \textsc{o2m} use 5000 and 16000 pieces, respectively. 

(ii) Syllabification (\textsc{Syl}) for Shipibo-Konibo is adapted from \citet{alva-oncevay-2017-spell}. The original method uses syllables to verify whether a word is composed by consistent syllables for spell-checking. In our experiments, when a word can not be syllabify-ed, we split it into characters for the \textsc{mono} setting, and we use the joint-BPE segmentation model for the \textsc{joint} and \textsc{o2m} settings. 

\paragraph{Model and training} We reproduce \citet{mager-etal-2022-bpe}'s settings, by using the fairseq toolkit \cite{ott-etal-2019-fairseq}, and a Transfomer model \cite{NIPS2017_7181} with smaller dimensions \cite{guzman-etal-2019-flores}. For the multilingual \textsc{o2m} setting, we use a sampling approach with 5 of temperature \cite{aharoni-etal-2019-massively}. See details in Appendix \ref{app:model}.

\paragraph{Evaluation} We use chrF \cite{popovic-2015-chrf} from \textsc{sacrebleu} \cite{post-2018-call}\footnote{chrF2+numchars.6+space.false+v.1.5.0.} 
and also perform a human evaluation of 100 samples per system (BPE and Syl), following the annotation protocol used in the AmericasNLP shared task \cite{mager-etal-2021-findings}.

\subsection{Results and discussion}

Table \ref{tab:mt-results} shows the translation performance in all settings, and we observe that syllables are statistically better than the BPE baseline when translating from Spanish into Shipibo-Konibo, but not in the other direction. This fact indicates that syllables support the decoding more than the encoding step of a language with a transparent orthography. Also, the \textsc{joint} setting reduces the gap between BPE and \textsc{Syl}, probably due to the shared roots between the two languages (i.e., loanwords from Spanish into Shipibo-Konibo). Furthermore, we note that the impact of syllables is not minimised in a multilingual system (\textsc{o2m}), where the performance for es$\rightarrow$shp has drastically improved, and the other language-pair (es$\rightarrow$en) retains a comparable result. 

Moreover, our \textsc{mono} experiments are comparable with the study of \citet{mager-etal-2022-bpe}, where they tested several unsupervised and supervised morphological segmentation methods against BPE for \MT in four polysynthetic languages (including Shipibo-Konibo). Our result with syllables in es$\rightarrow$shp outperforms all other approaches, such as LMVR \cite{ataman2017linguistically}, with a 38.99 chrF score. This indicates that syllables are a robust alternative to morphologically-aware methods when we are dealing with limited data and translating into a polysynthetic language. 

\begin{table}[]
\centering
\begin{tabular}{l|cc|p{0.45cm}p{0.45cm}}
                  & BPE                           & \textsc{Syl}                            & \small BPE                        & \small \textsc{Syl}               \\ \cline{2-5} 
                  & \multicolumn{2}{c|}{\textbf{es$\rightarrow$shp}}               & \multicolumn{2}{c}{\small \textbf{es$\rightarrow$en}} \\ \hline
\textsc{mono}     & 37.62\small$\pm$1.87          & \textbf{41.27}*\small$\pm$0.54 &                            &                   \\
\textsc{joint}    & 40.41\small$\pm$0.82          & \textbf{41.74}*\small$\pm$0.95 &                            &                   \\
\textsc{o2m} & 48.30                         & \textbf{51.25}*                & \small 53.99             & \small 53.85             \\ \hline
                  & \multicolumn{2}{c|}{\textbf{shp$\rightarrow$es}}               &                            &                   \\ \cline{1-3}
\textsc{mono}     & \textbf{33.37}\small$\pm$0.79 & 32.85*\small$\pm$1.22          &                            &                   \\
\textsc{joint}    & \textbf{34.55}\small$\pm$0.56 & 33.13*\small$\pm$0.75          &                            &                  
\end{tabular}

\caption{chrF scores in the test subsets. \textsc{Mono}: single BPE model (5k pieces) for each source and target. \textsc{Joint}: joint BPE model (5k) for both source and target. \textsc{o2m}: joint BPE model (16k) for \textsc{es}, \textsc{en} and \textsc{shp}. For the first two settings, we run three experiments and present the mean and standard deviation. The latter only has one run due to resource constraints, and we report es--en scores as a reference. Syllabification (\textsc{Syl}) is only applied on the Shipibo-Konibo side, and (*) indicates a p-value $\leq$ 0.05 against the BPE baseline.}
\label{tab:mt-results}
\end{table}

\subsection{Human evaluation}
We also conducted a small human evaluation of system outputs using a 5-points scale for the adequacy and fluency of the Spanish$\rightarrow$Shipibo-Konibo translation, which is the translation direction that benefited from the syllable segmentation. The annotation protocol and annotator's information is provided in Appendix \ref{app:annotation}.

\begin{figure}
    \centering
    \includegraphics[width=0.4925\linewidth,clip]{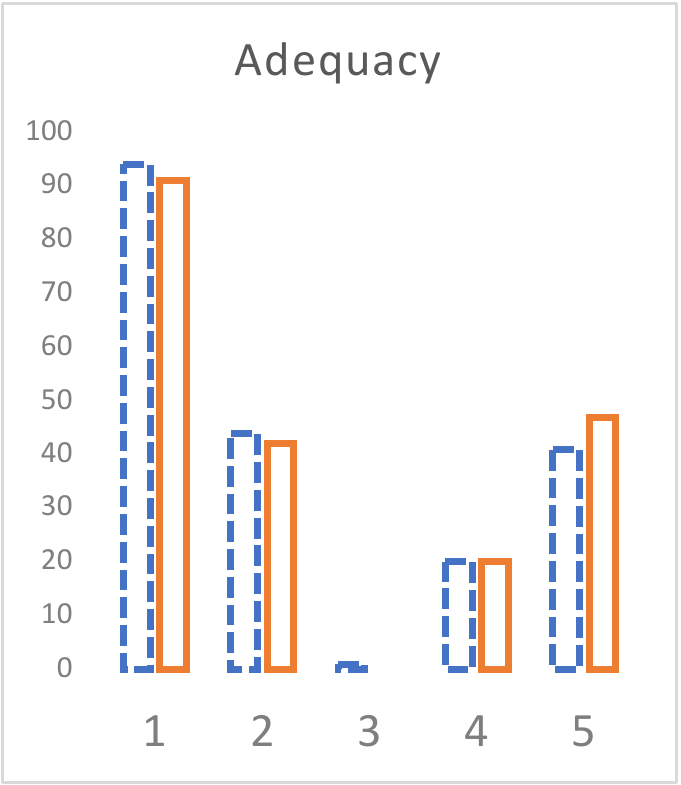}
    \includegraphics[width=0.4925\linewidth,clip]{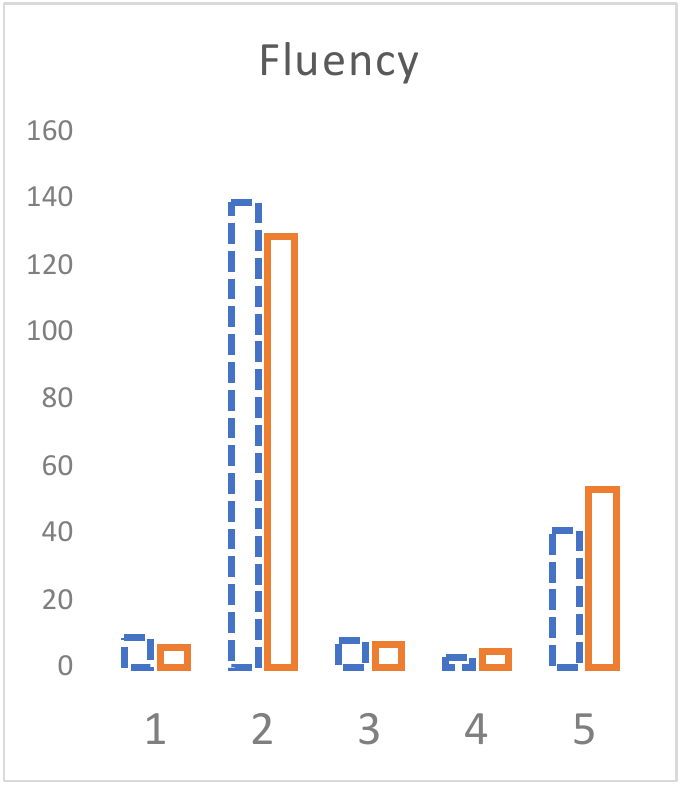}
    \caption{Adequacy and fluency scores (1-5) for 200 outputs of two approaches: BPE (dashed blue) and \textsc{Syl} (solid orange), from the best es$\rightarrow$shp \textsc{o2m} system.}
    \label{fig:human-eval}
\end{figure}

Figure \ref{fig:human-eval} shows the scores annotated for adequacy and fluency, where we compare BPE and \textsc{Syl} in the \textsc{o2m} setting, which obtained the best performance for both segmentation methods. 
We observe that the adequacy is very poor for both systems (1-2), but there is an advantage for \textsc{Syl} in the smaller batch of highest adequacy (5), with 3\% more of the total samples. Regarding fluency, both systems mostly obtain a low score (2), but there is a consistent advantage for \textsc{Syl} over BPE in the highest value (5), with 6.5\% more of the total samples. The differences are very small to determine whether a segmentation works better than the other from human judgement, but they are consistent with the automatic evaluation provided previously. A larger sample, an extra annotator, or more robust systems could aid to clarify other potential benefits.



\section{Limitations and opportunities} 

Syllables only cannot offer a universal solution to the subword segmentation problem for all languages, as the syllabification tools are language-dependent. Besides, the analysis should be extended to more scripts and morphological types. 
Furthermore, we do not encode any semantics in the syllable-vector space, with a few exceptions like in Korean \cite{choi-etal-2017-syllable}. 
Nevertheless, our results confirm that syllables are reliable for \LM and \MT, and building a syllable splitter might require less effort than annotating morphemes to train a robust supervised tool\footnote{For instance, the syllabification tool that we used for English is based on five general rules from: \url{https://www.howmanysyllables.com/divideintosyllables}. Their implementation should take less effort than annotating a UD treebank or building a Finite-State-Transducer for morphological analysis.}. 

Specifically for \MT, syllables could be useful when: (i) we are dealing with extremely low-resource data, which affects unsupervised word segmentation, (ii) we are translating into a language with a high synthesis, which has been observed as a factor that impacts on MT performance \cite{oncevay-etal-2022-quantifying}, and (iii) we are working with a language with a transparent orthography. 
This is the scenario for several languages from the Americas, where their writing systems have been recently  standardised for documentation and revitalisation purposes \cite{mager-etal-2018-challenges}, and some resources for \MT have been compiled \cite{mager-etal-2021-findings}.



\section{Conclusion}
We have proved that syllables are valuable for generation tasks such as: (i) Open-vocabulary \LM, where they behave positively even for languages with deep orthography, and overcome character and subword baselines. (ii) Low-resource and multilingual \MT, outperforming BPE pieces when we translate into a language with a transparent orthography and complex morphology (high synthesis), even when the language-pair is not related. 

\section*{Acknowledgements}
The first author acknowledges the support of NVIDIA Corporation with the donation of a Titan Xp GPU used for the study. The last author acknowledges the Max Planck Institute for Evolutionary Anthropology, Department of Linguistic and Cultural Evolution, for its support to the development of the Chana Field Station in the Amazonian region of Peru, and the support of CONCYTEC-ProCiencia, Peru, under the contract 183-2018-FONDECYT-BM-IADT-MU from the funding call E041-2018-01-BM.

\bibliography{custom}
\bibliographystyle{acl_natbib}

\newpage 
\appendix

\section{The Shipibo-Konibo language}
\label{app:shipibo}

Shipibo-Konibo (shp) is the largest and most vital language within the Pano language family. With more than 30,000 speakers, the Shipibo-Konibo are among the largest indigenous groups in Peru. Shipibo-Konibo people mainly live in the Peruvian Amazonia (in the regions of Ucayali, Loreto, Huánuco and Madre de Dios), but there are also large groups of Shipibo-Konibo people living in the Peruvian coast (particularly in Lima and Ica).

Shipibo-Konibo is a well-documented language, although the publicly available data on this language is rather small \cite{zariquiey-etal-2019-obsolescencia}. It has a complex morphology due to its high synthesis (high ratio of morphemes per word, mostly by suffixation) and it agglutinative nature. Its orthography can be considered transparent, because its alphabet was recently standardised by the Peruvian Government \cite{alva-oncevay-2017-spell}, and the datasets we are using in all experiments are provided with the most recent writing standard \cite{mager-etal-2021-findings}. Machine translation research on Shipibo-Konibo has focused on the development of new parallel corpora \cite{galarreta-etal-2017-corpus,montoya-etal-2019-continuous}, the application of multilingual models \cite{oncevay-2021-peru}, or the impact of morphological segmentation methods \cite{mager-etal-2022-bpe}. However, neither of them has focused on syllables as a unit for segmentation. For this study, we adapt the syllabification function proposed by \citet{alva-oncevay-2017-spell}, which was used for spell-checking.

\section{Dataset details}
\label{app:datasets}

Table \ref{tab:data-splits} shows the size of the training, validation and test splits for all the datasets used in the \LM task, while Table \ref{tab:data-mt} shows details of the Spanish--Shipibo-Konibo and Spanish--English parallel corpora used in the \MT task.

\begin{table}[h!]
    \centering
    \begin{tabular}{c|ccc}
                & train & dev & test \\ \hline
        es--shp & 13,102 & 587 & 1,030 \\
        es--en &  2,140,175 & 5,003 & 3,000
    \end{tabular}
    \caption{Total number of sentences in train, dev and test splits for the language-pairs used in the \MT experiments.}
    \label{tab:data-mt}
\end{table}

\begin{table*}[h!]
    \centering
    \small
    \setlength\tabcolsep{4pt}
    \begin{tabular}{l|rrr|rrr|rrr}
   & \multicolumn{3}{c|}{Train} & \multicolumn{3}{c|}{Valid} & \multicolumn{3}{c}{Test} \\ \hline
&  Word & Syl & Char & Word & Syl & Char & Word & Syl & Char \\ \hline \hline
en\textsubscript{w} & 2,089 & 4,894 & 10,902 & 218 & 505 & 1,157 & 246 & 568 & 1,304 \\ \hline
bg & 125 & 386 & 710 & 16 & 50 & 92 & 16 & 49 & 90 \\ 
ca & 436 & 1,123 & 2,341 & 59 & 152 & 317 & 61 & 157 & 327 \\ 
cs & 1,158 & 3,546 & 6,868 & 157 & 482 & 933 & 172 & 524 & 1,012 \\ 
da & 81 & 215 & 442 & 10 & 28 & 57 & 10 & 27 & 56 \\ 
de & 260 & 735 & 1,637 & 12 & 34 & 75 & 16 & 45 & 102 \\ 
en\textsubscript{} & 210 & 488 & 1,061 & 26 & 61 & 133 & 26 & 61 & 132 \\ 
es & 376 & 1,060 & 2,043 & 37 & 103 & 198 & 12 & 33 & 64 \\ 
fi & 165 & 595 & 1,224 & 19 & 67 & 137 & 21 & 76 & 155 \\ 
fr & 360 & 837 & 1,959 & 36 & 84 & 197 & 10 & 23 & 54 \\ 
hr & 154 & 484 & 930 & 20 & 62 & 119 & 23 & 75 & 145 \\ 
it & 263 & 762 & 1,504 & 11 & 32 & 64 & 10 & 28 & 57 \\ 
lv & 113 & 349 & 690 & 19 & 58 & 115 & 20 & 59 & 116 \\ 
nl & 187 & 488 & 1,074 & 12 & 30 & 66 & 11 & 31 & 68 \\ 
pl & 102 & 293 & 589 & 13 & 37 & 73 & 13 & 37 & 74 \\ 
pt & 192 & 551 & 1,040 & 10 & 29 & 54 & 9 & 27 & 51 \\ 
ro & 183 & 549 & 1,056 & 17 & 51 & 98 & 16 & 48 & 94 \\ 
ru & 867 & 2,707 & 5,411 & 118 & 364 & 722 & 117 & 360 & 717 \\ 
sk & 80 & 232 & 437 & 12 & 39 & 76 & 13 & 41 & 80 \\ 
tk & 38 & 126 & 242 & 10 & 33 & 63 & 10 & 33 & 64 \\ 
uk & 88 & 289 & 501 & 12 & 41 & 71 & 16 & 56 & 99 \\   \hline
shp & 43 & 141 & 398 & & & & & & \\ \hline
    \end{tabular}
    \caption{Total number of tokens (in thousands) at word, syllable and character-level for all the splits.}
    \label{tab:data-splits}
\end{table*}

\begin{figure*}[h!]
\begin{center}
\centering

\begin{subfigure}[t]{0.4\linewidth}
\includegraphics[width=\linewidth,clip]{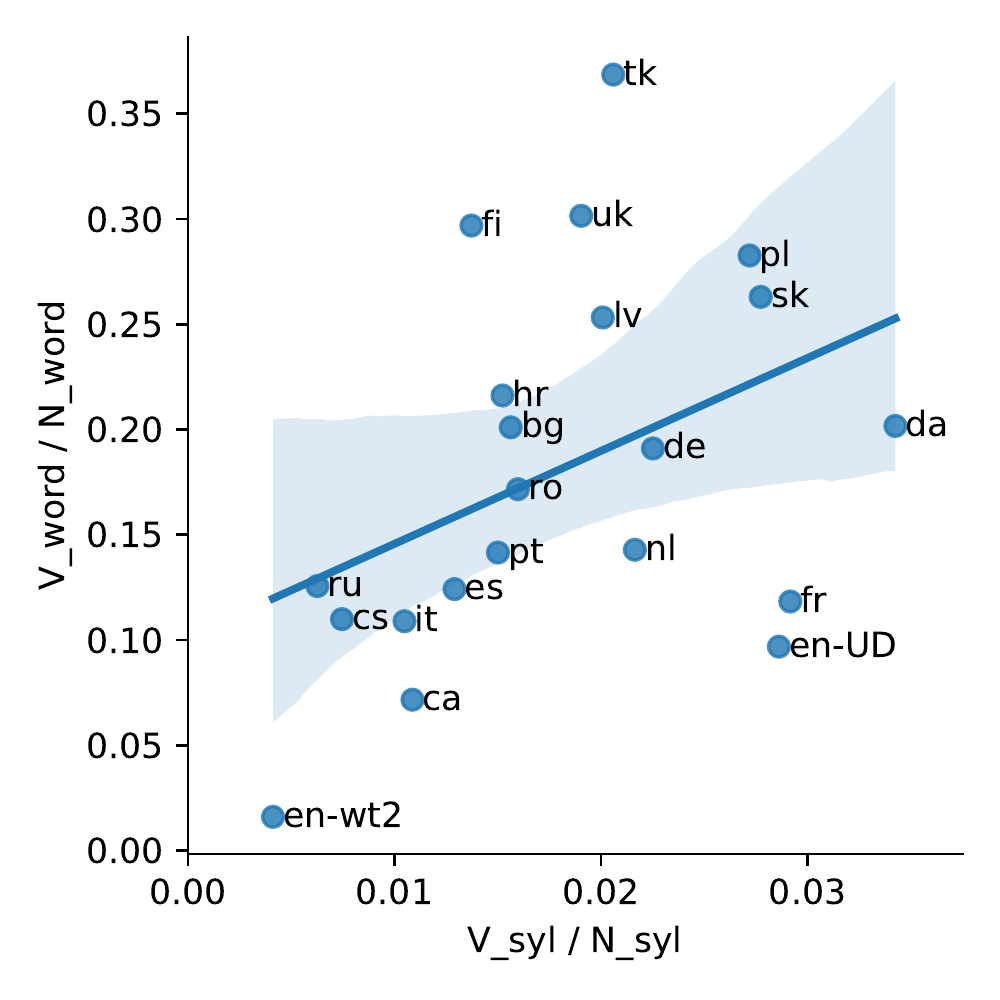}
\caption{V\textsubscript{syl}/N\textsubscript{syl} vs. V\textsubscript{word}/N\textsubscript{word}}
\label{fig:vocab-rate}
\end{subfigure}
\begin{subfigure}[t]{0.4\linewidth}
\includegraphics[width=\linewidth,clip]{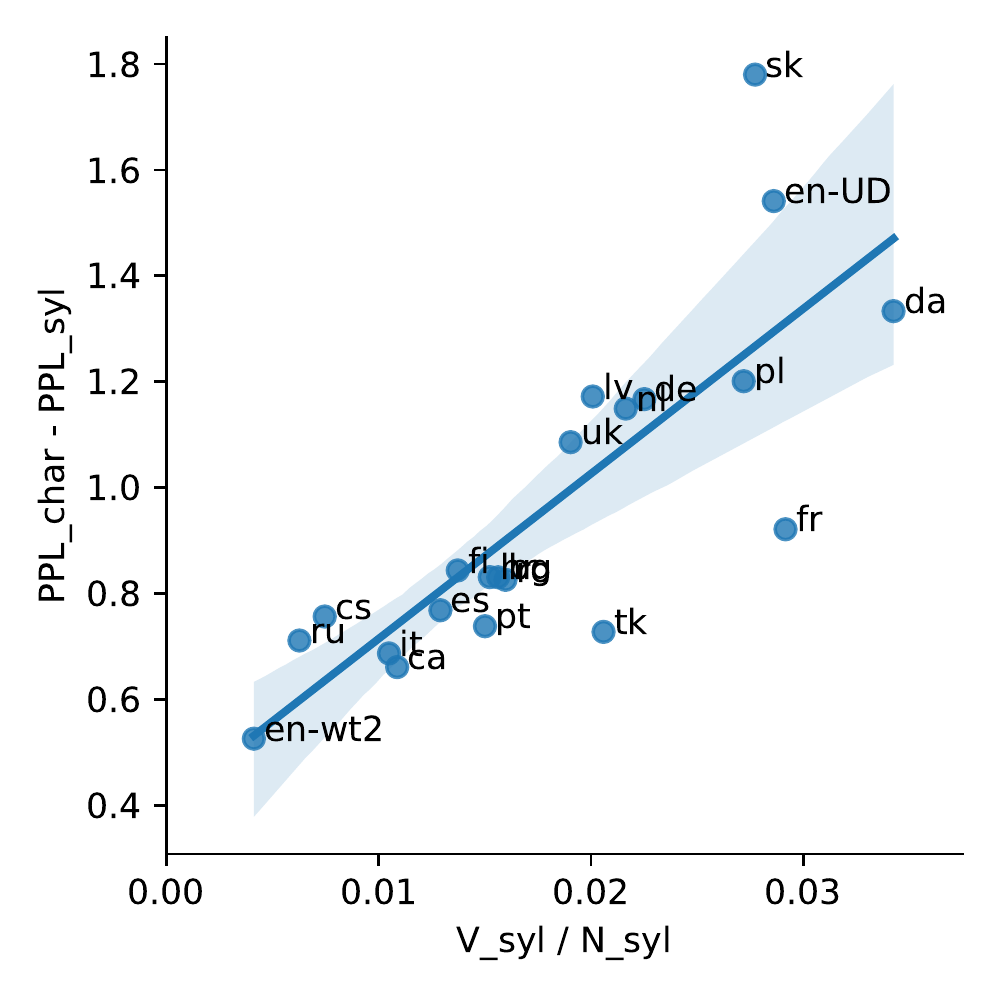}
\caption{V\textsubscript{syl}/N\textsubscript{syl} vs. $\Delta\operatorname{ppl}^c$\textsubscript{char-syl}}
\label{fig:ppl-diff}
\end{subfigure}

\caption{Left (a): Vocabulary growth rate of syllables (x-axis) versus words (y-axis). Right (b): Vocabulary growth rate of syllables (x-axis) versus the difference of $\operatorname{ppl}^c$ obtained by characters and syllables (y-axis).} 
\label{fig:regression-analysis}
\end{center}
\end{figure*}

\section{Segmentation details}
\label{app:segmentation}

\paragraph{Tools} We list the tools for rule-based syllabification and dictionary-based hyphenation:
\begin{itemize}
    \item English syllabification: Extracted from \url{https://www.howmanysyllables.com/}
    \item Spanish syllabification: \url{https://pypi.org/project/pylabeador/}
    \item Russian syllabification: \url{https://github.com/Koziev/rusyllab}
    \item Finnish syllabification: \url{https://github.com/tsnaomi/finnsyll}
    \item Turkish syllabification: \url{https://github.com/MeteHanC/turkishnlp}
    \item Shipibo-Konibo syllabification: \citet{alva-oncevay-2017-spell}
    \item Hyphenation: PyPhen (\url{https://pyphen.org/}), which is based on Hunspell dictionaries.
\end{itemize} 

\paragraph{Format} For syllables in the \LM task, 
we separate the subwords as: ``A @ syl la ble @ con tains @ a @ sin gle @ vow el @ u nit", where ``@'' is a special token that indicates the word boundary. We also evaluated syllables with a segmentation format like in \citet{sennrich-etal-2016-neural}: ``A syl@ la@ ble con@ tains a ...'', but we obtained lower performance in general. Whereas in the \MT task, we adopt the segmentation format used by SentencePiece~\cite{kudo-richardson-2018-sentencepiece} for syllables: ``\_A \_syl la ble \_con tains \_a \_sin gle \_vow el \_u nit''.

\section{Type/token ratio of syllables in LM}
\label{app:discusssion}

In Figure \ref{fig:vocab-rate}, we show a scatter plot of the token/type growth rate of syllables versus words for all languages and corpora. In other words, the ratio of syllable-types (syllabary or V\textsubscript{syl}) per total number of syllable-tokens (N\textsubscript{syl}) versus the type/token ratio of words (V\textsubscript{word}/N\textsubscript{word}) in the train set. 
The figure suggests at least a weak relationship, which agrees with the notion that a low word-vocabulary richness only requires a low syllabary richness for expressivity. Also, a richer vocabulary can use a richer syllabary or just longer words, so the distribution of the vocabulary richness could be larger.

We expected that the syllabary growth rate (V\textsubscript{syl}/N\textsubscript{syl}) for a low phonemic language like English would be relatively high, but wikitext-2 (en-wt2) is located in the bottom-left corner of the plot, probably caused by its large amount of word-tokens. However, we observed a large V\textsubscript{syl}/N\textsubscript{syl} for the English (en-UD) and French (fr) treebanks, despite their low V\textsubscript{word}/N\textsubscript{word} ratio, which is an expected pattern for deep orthographies. 

We also observe that languages with a more transparent orthography, like Czech (cs) or Finnish (fi), are located in the left side of the figure, whereas Turkish (tr) is around the middle section. 
Nevertheless, our study does not aim to analyse the relationship between the level of phonemic orthography with the V\textsubscript{syl}/N\textsubscript{syl} ratio. For that purpose, we might need an instrument to measure how deep or shallow a language orthography is \cite{marjou-2021-oteann, borgwaldt-etal-2005-onset, borleffs-etal-2017-measuring}, and a multi-parallel corpus for a more fair comparison.

Finally, in Figure \ref{fig:ppl-diff} we observe a stronger relationship of the syllable type/token ratio with the difference of \textsc{Char}'s $\operatorname{ppl}^c$ minus \textsc{Syl}'s $\operatorname{ppl}^c$. In other words, if our dataset possesses a rich syllabary, we are fairly approximating the amount of word-level tokens, which reduces the $\operatorname{ppl}^c$ gain.

\section{Model and Training}
\label{app:model}
\paragraph{LM} In contrast with the default settings, we use a smaller embedding size of 500 units for faster training. Additionally, we have 3 layers of depth, 1152 of hidden layer size and a dropout of 0.15. We train for 25 epochs with a batch size of 64, a learning rate of 0.002 and Adam optimiser \cite{kingma-ba-2015-adam} with default parameters. We fit the model using the one cycle policy and an early stopping of 4. We run our experiments in a NVIDIA Titan Xp.

\paragraph{MT} Similar to \citet{mager-etal-2022-bpe}, we use a small Transformer model for our low-resource \MT settings, following \citet{guzman-etal-2019-flores}: ``with 5 encoder and 5 decoder layers, where the number of attention heads, embedding dimension and inner-layer dimension are 2, 512 and 2048, respectively''. For the pairwise systems, we train up to 100 epochs with an early stopping policy of 5 (validating every 5 epochs), whereas for the multilingual systems we train up to 30 epochs. For all the experiments, we use 4 NVIDIA GeForce GTX 1080 Ti GPUs.

\section{Human evaluation}
\label{app:annotation}

\subsection{Annotation protocol} 

Adapted and summarised from the AmericasNLP shared task \cite{mager-etal-2021-findings}: The expert received the source sentence in Spanish, the reference in Shipibo-Konibo, and an anonymized system output, which includes the baseline (BPE) and our syllable-based system (\textsc{Syl}). The expert received only 200 samples (per system, same entries) that were randomly selected and shuffled. They were asked to annotate \textbf{Adequacy} (Does the output sentence express the meaning of the reference?) from 1 to 5 (extremely bad, bad, neutral, sufficiently good, excellent), and \textbf{Fluency} (Is the output sentence easily readable and looks like a human-produced text?) from 1 to 5 as well.




\subsection{About the annotator}

The annotator is a native speaker of Shipibo-Konibo, a certified and professional translator, and a bilingual teacher in Peru. The annotator has experience in translating corpus for \MT research, and performing human evaluation for Spanish--Shipibo-Konibo. This expertise is almost unique for Shipibo-Konibo, and we could not identify a second annotator with the same expertise to obtain inter-annotation agreement.

\end{document}